\pdfoutput=1

\documentclass[11pt]{article}

\usepackage[final]{acl}

\usepackage{times}
\usepackage{latexsym}

\usepackage[T1]{fontenc}

\usepackage[utf8]{inputenc}

\usepackage{microtype}

\usepackage{inconsolata}

\usepackage{times}
\usepackage{latexsym}
\usepackage{algorithm}
\usepackage{algpseudocode}
\usepackage{amsmath}
\usepackage{amssymb}
\usepackage{amsmath}
\usepackage{arydshln}
\usepackage{ascii}
\usepackage{bm}
\usepackage{booktabs}
\usepackage{colortbl}
\usepackage{color}
\usepackage{CJK}
\usepackage{dsfont}
\usepackage{enumitem}
\usepackage{forest}
\usepackage{graphics}
\usepackage{graphicx}
\usepackage{latexsym}
\usepackage{multirow}
\usepackage{mwe}
\usepackage{makecell}
\usepackage{microtype}
\usepackage{subfig}
\usepackage{subcaption}
\usepackage{times}
\usepackage{tikz}
\usepackage{tabularx}
\usepackage{url}
\usepackage{wrapfig}
\usepackage{tcolorbox}
\usepackage{xcolor}
\usepackage{hyperref}

\definecolor{comments}{RGB}{250, 0, 0}

\title{\includegraphics[width=1.0em]{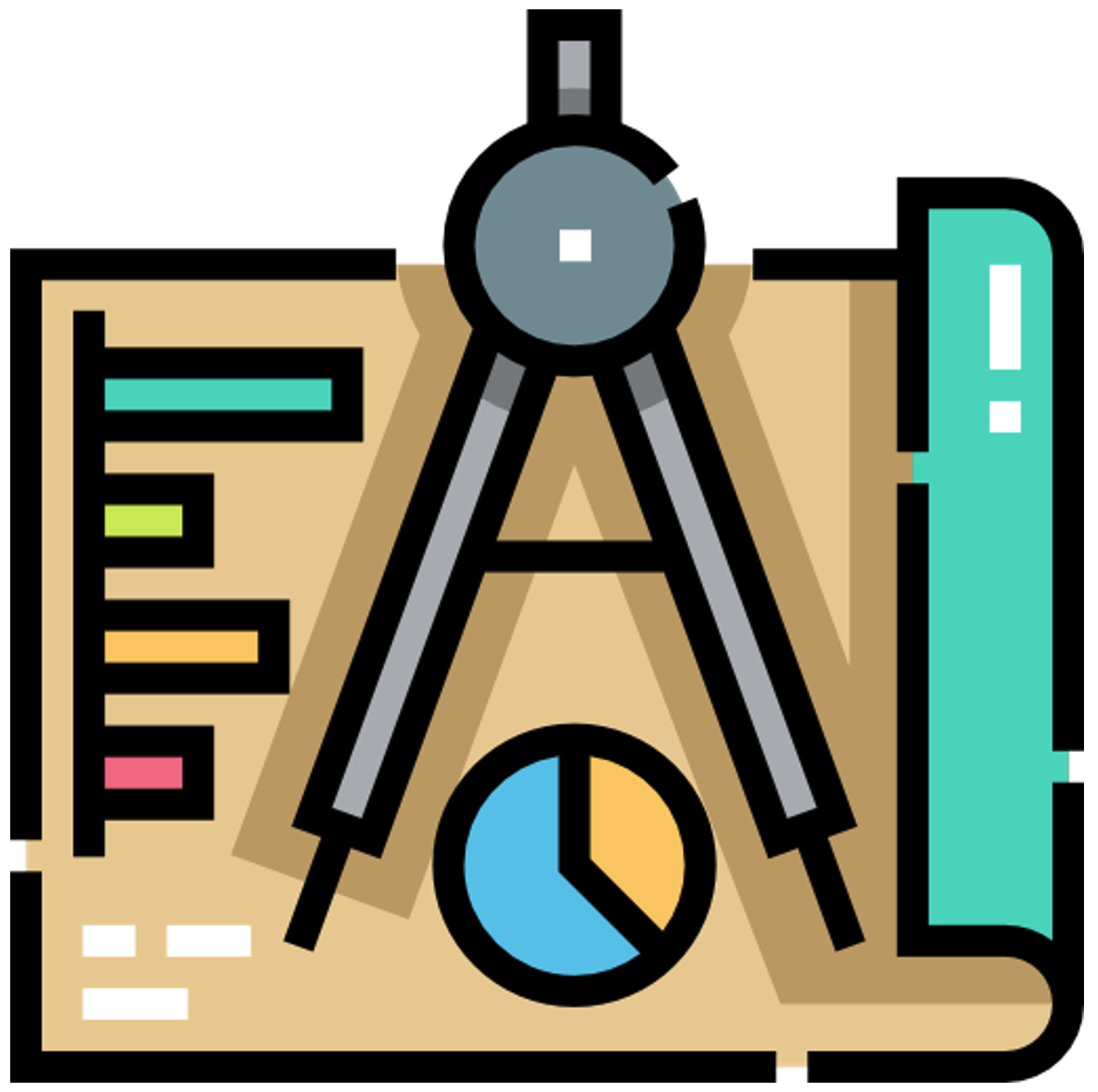}\ MatPlotAgent: Method and Evaluation for\\
LLM-Based Agentic Scientific Data Visualization}

\author{Zhiyu Yang$^{*2}$ \
  Zihan Zhou\thanks{\ \ Equal contribution.}$^{3}$ \
  Shuo Wang$^{\dag1}$ \
  Xin Cong$^{1}$ \\
  \textbf{Xu Han}$^{1}$ \
  \textbf{Yukun Yan}$^{1}$ \
  \textbf{Zhenghao Liu}$^{4}$ \
  \textbf{Zhixing Tan}$^{5}$ \\
  \textbf{Pengyuan Liu}$^{2}$ \
  \textbf{Dong Yu}$^{2}$ \
  \textbf{Zhiyuan Liu}\thanks{\ \ Corresponding authors.}$^{1}$ \
  \textbf{Xiaodong Shi}$^{3}$
  \textbf{Maosong Sun}$^{1}$ \\
  $^1$Tsinghua University \ $^2$Beijing Language and Culture University \ $^3$Xiamen University\\
  $^4$Northeastern University, China \ $^5$Zhongguancun Laboratory, Beijing, China\\
  }

\begin{document}
\maketitle
\begin{abstract}
Scientific data visualization plays a crucial role in research by enabling the direct display of complex information and assisting researchers in identifying implicit patterns. Despite its importance, the use of Large Language Models (LLMs) for scientific data visualization remains rather unexplored. In this study, we introduce MatPlotAgent, an efficient model-agnostic LLM agent framework designed to automate scientific data visualization tasks. Leveraging the capabilities of both code LLMs and multi-modal LLMs, MatPlotAgent consists of three core modules: query understanding, code generation with iterative debugging, and a visual feedback mechanism for error correction.
To address the lack of benchmarks in this field, we present MatPlotBench, a high-quality benchmark consisting of 100 human-verified test cases. Additionally, we introduce a scoring approach that utilizes GPT-4V for automatic evaluation. Experimental results demonstrate that MatPlotAgent can improve the performance of various LLMs, including both commercial and open-source models. Furthermore, the proposed evaluation method shows a strong correlation with human-annotated scores.\footnote{\ \ MatPlotAgent and MatPlotBench are be publicly available at \url{https://github.com/thunlp/MatPlotAgent}.}
\end{abstract}

\section{Introduction}

{\em A picture is worth a thousand words.} Data visualization is an essential process in scientific research, facilitating the more direct conveyance of complex information and aiding researchers in uncovering implicit patterns. There are many advanced toolkits, such as Matplotlib\footnote{\url{https://matplotlib.org}} and Origin\footnote{\url{https://www.originlab.com}}, that can help researchers plot various types of figures for complex data distributions. However, transforming raw data into informative and easy-to-understand visualizations is still time-consuming and labor-intensive. Before the invention of large language models (LLMs)~\cite{Achiam2023GPT4TR}, automating this process with AI models is almost impossible.

\begin{figure*}[t]
    \centering
    \includegraphics[width=0.99\linewidth]{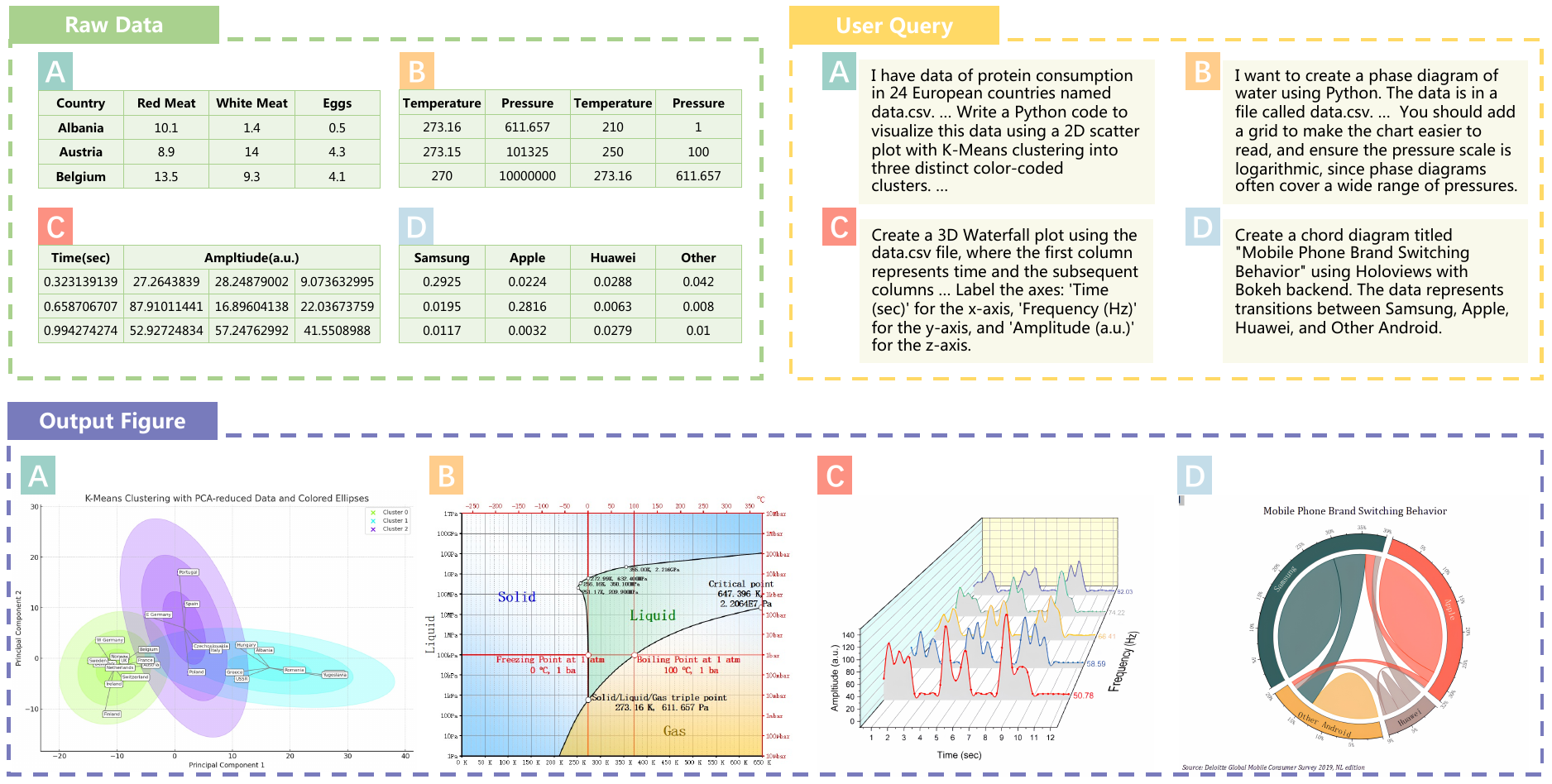}
    \caption{Examples in the proposed MatPlotBench. Given the raw data and user queries, the AI agent is expected to generate a figure accordingly. We only display partial raw data and user queries due to space limitations.}
    \label{fig:Task_description}
\end{figure*}

With large-scale parameters and extensive training data, LLMs have demonstrated remarkable capabilities in a wide range of complex tasks, including reasoning~\cite{NEURIPS2022_9d560961_CoT,NEURIPS2022_zero-shot-cot,yao2023treeToT}, mathematics~\cite{yu2024metamath,luo2023wizardmath,azerbayev2024llemma,shao2024deepseekmath} and coding~\cite{rozière2024code, luo2023wizardcoder, guo2024deepseekcoder, wei2023magicoder}. This breakthrough has unlocked new opportunities for utilizing LLMs as autonomous agents in a diverse range of practical scenarios, such as web browsing~\cite{Nakano2021WebGPTBQ,yao_webshop_2022,qin-etal-2023-webcpm,zhou2023webarena,deng2023mindweb, yao2023react,xie2023openagents}, social simulations~\cite{Generative_agents,xu2023exploringWerewolf,chen2024agentverse,Wang2023VoyagerAO}, tool utilization~\cite{qin2024toolllm,schick2023toolformer,liu2024agentbench,li2023camel,lu2023chameleon,qian-etal-2023-creator,shinn2023reflexion}, and software development~\cite{qian2023ChatDev}. Using LLMs to enhance human productivity in specialized areas is now a key research focus with great potential.

Recent advancements in LLM-based agents inspire us to explore the utilization of LLMs for scientific data visualization, a realm that remains rather unexplored in existing studies.
A closely related line of research is text-to-image generation~\cite{ramesh2021zeroshotDalle,saharia2022photorealisticImagen}, where diffusion models~\cite{rombach2022highresolutionStableDiffusion} have shown great potential in generating various types of images.
However, existing text-to-image generation methods predominantly focus on artistic expression, potentially misaligning with the needs of scientific data visualization, where clarity and precision in conveying information are the most important principles. This work aims to automatically generate figures with precise information.

We propose leveraging modern code LLMs and multi-modal LLMs to develop scientific data visualization agents that can significantly enhance human efficiency. The resulting \textbf{MatPlotAgent}\footnote{This name is in homage to the well-known Matplotlib.} is comprised of three modules: (1) {\em the query understanding} that can thoroughly understand user-provided requirements; (2) {\em the code generation module} with iterative debugging capabilities that use code to precisely preprocess raw data and generate figures; and (3) {\em the visual feedback module} that possesses visual perceptual abilities to find errors in the plotted draft and provide visual feedback to the code generation module to rectify the errors.
Our method is model-agnostic, which can be driven with any code LLMs and multi-modal LLMs. Through experiments, we find MatPlotAgent can work with both closed-source LLMs (e.g., GPT-4~\cite{Achiam2023GPT4TR}) and open-source LLMs (e.g., Magicoder~\cite{wei2023magicoder}).

Another critical challenge in the field of automatic scientific data visualization is the absence of benchmarks for evaluation purposes. To address this issue, we introduce a meticulously crafted benchmark called \textbf{MatPlotBench} to quantitatively evaluate the approaches involved. Specifically, MatPlotBench contains 100 carefully hand-crafted test examples, each of which contains a user query, the corresponding input data, and a ground-truth figure verified by human experts. We believe that high-quality test sets play a crucial role in driving advancements in the field.

To facilitate automatic quantitative evaluation, we also design a scoring mechanism based on GPT-4V~\cite{Achiam2023GPT4TR}, which is one of the strongest multi-modal LLMs that can effectively understand text and figures. 
Specifically, GPT-4V is prompted to produce a score between 0 and 100 based on the ground-truth figure and the one generated by AI models. Additionally, we conduct human evaluation and estimate the correlation coefficient between human-annotated scores and the automatically calculated scores. The results reveal a strong correlation between the automatic score and the human-annotated score, thus affirming the reliability of the scoring mechanism.
In summary, our contribution can be listed as follows:
\begin{itemize}
    \item We introduce MatPlotBench to enable automatic quantitative evaluation of AI methods designed for scientific data visualization. Through comparison with human evaluation, we observe that MatPlotBench can effectively capture the performance of AI approaches in this cutting-edge task.
    \item We propose an effective and generalizable LLM agent framework, MatPlotAgent, that can improve the performance of a wide range of LLMs based on the newly proposed visual feedback mechanism.
\end{itemize}

\section{Task Description}

We first introduce the scientific data visualization task investigated in this work. Given a user query $\mathbf{x}$ described in text and the corresponding data $\mathcal{D}$, the AI system is expected to output a figure $V$ that can satisfy the user's demand:
\begin{equation}
    V = f(\mathbf{x}, \mathcal{D}),
\end{equation}
where $f$ denotes the involved AI system that can be either an LLM or an LLM-based agent.

Specifically, $\mathbf{x}$ specifies the visualization requirements, encompassing the visualization type, data to plot, structural or spatial requirements for individual elements or the entire plot, and aesthetic preferences. $\mathcal{D}$ represents the data, a collection of data points $\left\{ d_1, \cdots, d_n \right\}$ whether specified by the user or stored in the external data file. Figure~\ref{fig:Task_description} provides some examples for this task.

\section{MatPlotBench}

Automatic evaluation is important in AI tasks as it enables researchers to efficiently assess the performance of various methods, thereby guiding the development of the field. While the DS-1000 benchmark~\cite{Lai2023DS1000} includes coding problems about Matplotlib, the solutions' average length is merely three lines, rendering them too simplistic to gauge the proficiency of contemporary AI agents in tackling practical challenges. Therefore, we propose to construct MatPlotBench with complex data visualization problems that are more close to real-world scenarios. We will illustrate the data collection process in Section~\ref{sec:data-collection} and then explain the scoring mechanism in Section~\ref{sec:scoring}.

\subsection{Data Collection}
\label{sec:data-collection}

\paragraph{Principles}
To enhance the quality of MatPlotBench, we adhere to the following principles for data collection: (1) {\em Covering diverse types}: encompassing a broad range of plot types, including not only the most commonly used but also rare but useful ones; (2) {\em Containing representative instances}: ensuring that the test examples reflect the representative features of scientific data visualization, such as varying data complexity; and (3) {\em Balancing easy and challenging problems}: including problems of varying levels of difficulty in the benchmark.

\paragraph{Selecting Original Examples} 
In accordance with the principles outlined above, we first select some original examples from reputable online scientific data visualization forums. These examples are carefully selected from the Matplotlib Gallery and OriginLab GraphGallery, encompassing diverse and representative instances with varying levels of difficulty. 
Specifically, we select 1 or 2 examples from every section in the Matplotlib Gallery, covering bars, lines, markers, pie charts, polar plots, contour plots, statistics plots, 3D plots, text annotations, radar charts, shapes, scales, axes, spines, subplots, and so on. We also seek more advanced test examples from the OriginLab GraphGallery, focusing on those that are more aesthetically appealing or complex, such as Sankey diagrams, sunburst charts, radial plots, chord diagrams, streamplots, and others. Finally, 75 original examples come from the Matplotlib Gallery and the 25 other original examples come from the OriginLab GraphGallery. Subsequently, these examples undergo several modifications to become the final test cases in MatPlotBench. 

\paragraph{Preliminary Query Generation}

Based on the selected original examples, we use LLMs to generate preliminary queries, which are then revised by humans. For original examples from the Matplotlib Gallery, we use GPT-4 to convert the code in each original example into preliminary queries. For the examples from the OriginLab GraphGallery, there are only images. We thus use GPT-4V to convert each image into a preliminary query.

\paragraph{Data Replacement}

Based on these preliminary queries, we begin data replacement for examples from the Matplotlib Gallery due to the observed phenomenon of memorization by GPT-4. In this process, we replace the original data points with newly generated ones, while keeping other factors such as the plot type unchanged. For examples from OriginLab, we find that the data is inherently complex, and even GPT-4 does not exhibit memorization with these examples. As a result, we only perform data replacement for Matplotlib examples.

\paragraph{Human Modification} After completing the data replacement process, we engage human annotators to refine the preliminary queries. These annotators are tasked with correcting errors, eliminating ambiguity, and adding any omitted essential information. Each annotator involved has a minimum of three years of experience in coding and NLP. Furthermore, each query undergoes refinement by two independent human annotators.

\paragraph{Updating Ground-Truth Figures} After obtaining the human-annotated queries, as the data in Matplotlib examples are altered, we cannot directly use the images in the original example as the ground truth. To this end, we manually wrote code to plot the ground truth for the Matplotlib examples. For examples from OriginLab, as the data remains unaltered, we extract the images from their website to serve as the ground truth.

\paragraph{Human Verification}

After obtaining the queries and their corresponding ground truths, we performed a final round of manual verification. Three NLP researchers were asked to conduct this verification. In this turn, the focus is mainly on checking whether the user queries and the ground truths are well aligned. The researchers meticulously checked each element in the ground truth image and looked for their corresponding descriptions in the user query. Ill-described elements and those missing clarifications are corrected. Redundant and incorrect descriptions are removed. This process results in 100 high-quality (\textit{query, raw data, ground-truth figure}) triples, which comprise our final benchmark.

\subsection{Automatic Quantitative Evaluation}
\label{sec:scoring}

To ease the burden of manual evaluation and broaden the applicability of our benchmark for research purposes, we suggest employing GPT-4V, a cutting-edge multi-modal LLM, to conduct automatic evaluations on our proposed benchmark.
We carefully prompt GPT-4V to give a score from 0 to 100 on model-generated visualizations using the corresponding ground truths as the reference. The prompt is shown in Figure~\ref{fig:4v_prompt} in Appendix.

\paragraph{Correlation with Human Evaluation}
To assess the reliability of GPT-4V as an automatic evaluator for scientific visualizations, we calculate the correlation between the automatic scores and human-evaluated scores. Specifically, we employ GPT-3.5 and GPT-4 to generate figures on MatPlotBench, and then conduct both automatic and human evaluation for the generated figures. For each model, we iteratively sample a subset that consists of $n$ examples from the total benchmark, and then calculate the average score of both automatic and human evaluation. This process repeats $k$ times and we get $k$ data points for each type of evaluation, which can be represented by $\mathcal{A} = \{ a_1, \cdots, a_k \}$ and $\mathcal{H} = \{ h_1, \cdots, h_k \}$. $a_i$ denotes the average automatic score on the $i$-th randomly sampled subset, and $h_i$ represents the average human-evaluated score in the same subset. $n$ and $k$ are set to 25 and 100, respectively.

\begin{figure}[t]
    \centering
    \includegraphics[width=0.9\linewidth]{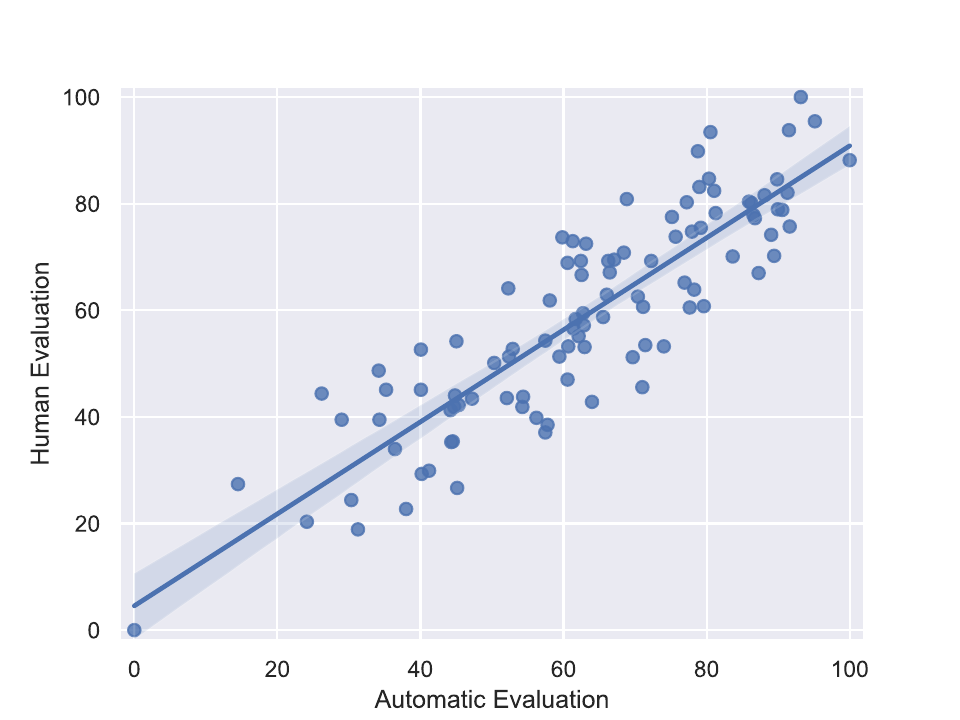}
    \caption{Correlation between the proposed automatic evaluation mechanism and human evaluation.}
    \label{fig:corr}
\end{figure}

We utilize the statistical functions provided by \texttt{scipy}\footnote{\url{https://docs.scipy.org/doc/scipy/reference/stats.html}} to compute the Pearson correlation coefficient $r$ and the corresponding p-value $p$. For GPT-4, we obtain $r$=0.876 and $p$=7.41e-33, while for GPT-3.5, the values are $r$=0.836 and $p$=2.67e-27. Figure~\ref{fig:corr} shows the data points for GPT-4. Given that $r>0.8$ and $p<$0.05, we conclude that the automatic evaluation scores are strongly correlated with human evaluation results. This demonstrates the reliability of the proposed scoring mechanism in assessing the quality of model-generated figures on MatPlotBench.

\section{MatPlotAgent}
\label{sec:agent}
\begin{figure*}[t]
    \centering
    \includegraphics[width=0.99\linewidth]{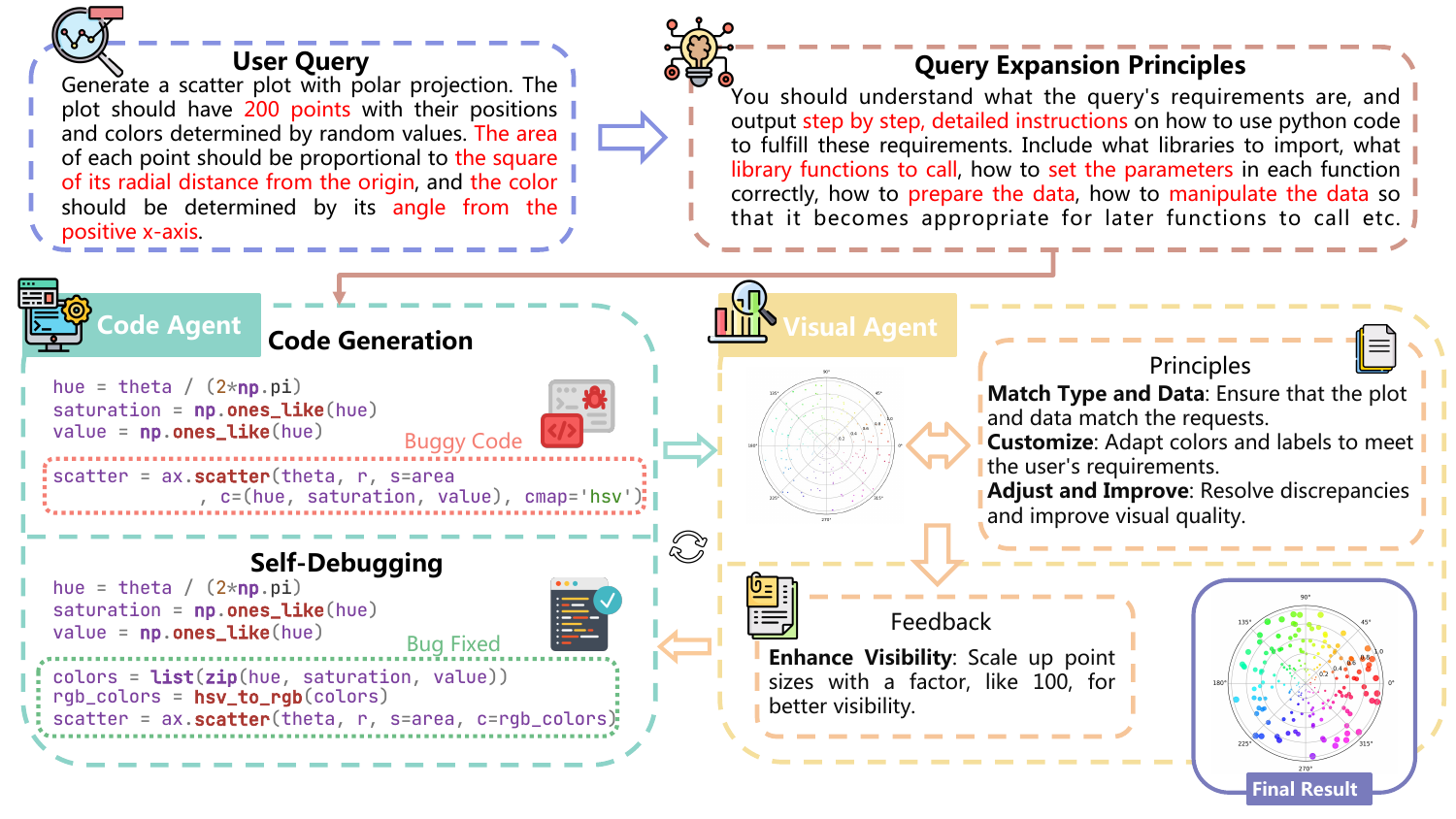}
    \caption{Workflow of MatPlotAgent: The query expansion module converts the user query into detailed multi-step instructions. These instructions are then passed to the code agent, which generates the plotting code. The visual agent provides informative feedback based on the current draft, guiding the refinement of the figure.}
    \label{fig:workflow}
\end{figure*}

To improve the capabilities of LLMs for scientific data visualization, we propose an agentic framework that mimics the plotting process of human experts. The proposed MatPlotAgent is comprised of three modules, including the query expansion module, the code agent, and the visual agent. Figure~\ref{fig:workflow} illustrates the workflow of MatPlotAgent.

\subsection{Query Expansion}

The query expansion module interprets and refines the user query, converting the high-level requirements into a sequence of explicit and detailed instructions that are easy for LLMs to follow. This module can also be viewed as a planning module, creating an overall plan before generating the figure. Specifically, this module is based on the involved code LLM, which is prompted to give detailed instructions on how to use code to fulfill the requirement specified by the user, including what libraries to import, what library functions to call, how to set the parameters in each function correctly, how to prepare the data, how to manipulate the data, and so on.

\subsection{Code Agent}

The code agent is the core component in MatPlotAgent, responsible for generating the code to plot figures.
Given detailed instructions from the query expansion module, the code agent first generates the code using appropriate libraries and functions. To improve the success rate of the generated code, we also employ the self-debugging mechanism~\cite{chen2024teaching}, which helps the involved code LLM iteratively identify and correct bugs in the code. To prevent an infinite loop, we set the maximum iterations of self-debugging to 3.

Similar to humans, who need to repeatedly refine the figure based on current drafts, we also introduce a visual feedback mechanism. This mechanism employs multi-modal LLMs to provide suggestions to improve the figure and better fulfill the user's queries. These suggestions, which we call {\em visual feedback}, are then provided to the code agent to further improve the code.
Our experiments in Section~\ref{sec:mainres} demonstrate that MatPlotAgent is compatible with several modern code LLMs, including both some well-known closed-source models and some open-source models.

\begin{table*}[ht]
\centering
\begin{tabular}{l c rr rr}
\toprule
\multirow{2}{*}{\textbf{Model}} & \multicolumn{1}{c}{\textbf{Direct}} & \multicolumn{2}{c}{\bf Zero-Shot} & \multicolumn{2}{c}{\textbf{MatPlotAgent}} \\
& \bf Decod. & \multicolumn{2}{c}{\bf CoT} & \multicolumn{2}{c}{\textbf{w/ GPT-4V}}\\
\cmidrule(lr){1-1} \cmidrule(lr){2-2} \cmidrule(lr){3-4} \cmidrule(lr){5-6}
GPT-4 & 48.86 & 45.42 & {\cellcolor{blue!22}} \small $-$3.44 & 61.16 & {\cellcolor{red!22}} \small $+$12.30\\
GPT-3.5 & 38.03 & 37.14 & {\cellcolor{blue!22}} \small $-$0.89 & 47.51 & {\cellcolor{red!22}} \small $+$9.48 \\
\cmidrule(lr){1-1} \cmidrule(lr){2-2} \cmidrule(lr){3-4} \cmidrule(lr){5-6}
Magicoder-S-DS-6.7B~\cite{wei2023magicoder} & 38.49 & 37.95 & {\cellcolor{blue!22}} \small $-$0.54 & 51.70 & {\cellcolor{red!22}} \small $+$13.21 \\
Deepseek-coder-6.7B-instruct~\cite{guo2024deepseekcoder} & 31.53 & 29.16 & {\cellcolor{blue!22}} \small $-$2.37 & 39.45 & {\cellcolor{red!22}} \small $+$7.92 \\
CodeLlama-34B-Instruct~\cite{rozière2024code} & 16.54 & 12.40 & {\cellcolor{blue!22}} \small $-$4.14 & 14.18 & {\cellcolor{blue!22}} \small $-$2.36 \\ 
Deepseek-coder-33B-instruct~\cite{guo2024deepseekcoder}  & 30.88 & 36.10 & {\cellcolor{red!22}} \small $+$5.22 & 32.18 & {\cellcolor{red!22}} \small $+$1.30 \\
WizardCoder-Python-33B-V1.1~\cite{luo2023wizardcoder}  & 36.94 & 35.81 & {\cellcolor{blue!22}} \small $-$1.13 & 45.96 & {\cellcolor{red!22}} \small $+$9.02 \\
\bottomrule
\end{tabular}
\caption{
Performance of different LLMs on MatPlotBench. For each model, improvements over the direct decoding are highlighted in {\color{red!50} red}, while results worse than that of the direct decoding are highlighted in {\color{blue!50} blue}.
}
\label{tab:LLM_score}
\end{table*}

\begin{table}[ht]
\centering
\begin{tabular}{l c c r r c}
\toprule
\multirow{2}{*}{\textbf{Model}} & \textbf{Direct}  & \multicolumn{4}{c}{\textbf{MatPlotAgent}} \\
& \bf Decod.  &  \multicolumn{4}{c}{\bf w/ Gemini Pro Vision } \\
\cmidrule(lr){1-1} \cmidrule(lr){2-2} \cmidrule(lr){3-6}
GPT-4 & 48.86 & &  56.73 & {\cellcolor{red!22}} \small $+$7.87 & \\
GPT-3.5 & 38.03 & &  43.48 & {\cellcolor{red!22}} \small $+$5.45 & \\
\bottomrule
\end{tabular}
\caption{
Performance of GPT-4 and GPT-3.5 using Gemini Pro Vision as visual agent on MatPlotBench.
}
\label{tab:LLM_Gemini_vf_score}
\end{table}

\subsection{Visual Agent}

The major difference between MatPlotAgent and previous LLM-based coding agents~\cite{qian2023ChatDev,chen2024teaching} is that we take the visual signal into account, which is important in scientific data visualization. 
Some errors or weaknesses may be difficult to identify in the code but become apparent when observing the output figure through ``eyes''. The visual agent is the ``eyes'' for MatPlotAgent, while the aforementioned code agent acts as the ``hands'' for MatPlotAgent.

Specifically, the visual agent is powered by multi-modal LLMs.
We introduce several guiding principles for the visual agent, including verifying whether the figure aligns with the provided data, and enhancing the colors or labels to improve the figure's informativeness. Based on the principles, the user query, and the current draft of the figure, the visual agent generates some suggestions to refine to figure. These suggestions serve as feedback for the code agent to refine the code.
Experimental results in Section~\ref{sec:ablation} show that our visual feedback mechanism can significantly improve the quality of the plotted figures.

\section{Experiments}
\subsection{Setup}

\paragraph{Models} Since the proposed MatPlotAgent is model-agnostic, we can employ various LLMs in this framework. The code LLMs we use in our experiments include GPT-4, GPT-3.5, Magicoder-S-DS-6.7B~\cite{wei2023magicoder}, Deepseek-coder-6.7B-instruct~\cite{guo2024deepseekcoder}, Deepseek-coder-33B-instruct~\cite{guo2024deepseekcoder}, WizardCoder-Python-33B-V1.1~\cite{luo2023wizardcoder}, and CodeLlama-34B-Instruct~\cite{rozière2024code}. The decoding temperature is set to 0.0 for all the involved code LLMs. For GPT-4 and GPT-3.5, we use the API provided by OpenAI\footnote{\url{https://openai.com/product}}. For the other five open-source LLMs, we use \texttt{vLLM}~\cite{kwon2023efficient} for model inference.
For the visual agent, we utilize GPT-4V~\cite{Achiam2023GPT4TR} and Gemini Pro Vision \cite{geminiteam2023gemini}, two representative multi-modal LLMs. We leave the exploration of using open-source multi-modal LLMs to power the visual agent for future work.

\paragraph{Evaluation} We evaluate the involved methods on MatPlotBench, using the proposed automatic scoring mechanism that is shown reliable in Section~\ref{sec:scoring}.
For each code LLM, we evaluate its performance in three ways:
\begin{itemize}
    \item Direct decoding: given the query, the model directly generates the plotting code.
    \item Zero-Shot Chain-of-thought~\cite{NEURIPS2022_8bb0d291}: the model is prompted to inference with the zero-shot CoT mechanism.
    \item MatPlotAgent: the model is equipped with the proposed MatPlotAgent framework, driving the query expansion module and the code agent, as illustrated in Section~\ref{sec:agent}.
\end{itemize}

\begin{table*}[t]
\centering
\begin{tabular}{l ccc}
\toprule
\multirow{2}{*}{\textbf{Model}} & \multicolumn{3}{c}{\textbf{Accuracy of Code Execution Results (\%)}} \\
\cmidrule(lr){2-4}
 & \textbf{Visualization-Hard} & \textbf{Visualization-Easy} & \bf Average \\ \midrule
GPT-4 & 66.7 & 60.8 & 63.8 \\ 
\ + MatPlotAgent & \bf 72.6 & \bf 68.4 & \bf 70.5 \\
\ \ \ \  w/o Visual Feedback & 66.7 & 65.8 & 66.3 \\ \bottomrule
\end{tabular}
\caption{Effect of MatPlotAgent on the visualization subset of the Qwen-Agent Code Interpreter benchmark.}
\label{tab:qwen_benchmark_vis_study_results}
\end{table*}

\subsection{Main Results}
\label{sec:mainres}
Table~\ref{tab:LLM_score} presents the results of different methods on the scientific data visualization task. In the direct decoding setting, GPT-4 achieves the highest score of 48.86. Surprisingly, the open-source model Magicoder-S-DS-6.7B~\cite{wei2023magicoder} achieves the second-best performance, surpassing models with substantially larger parameter sizes, such as WizardCoder-Python-33B-V1.1.

The results also suggest that the zero-shot CoT mechanism does not effectively enhance the performance of many recent code LLMs. Zero-shot CoT only improves the results of Deepseek-coder-33B-instruct~\cite{guo2024deepseekcoder} from 30.88 to 36.10. Conversely, for other models, implementing zero-shot CoT results in poorer performance. For example, when zero-shot CoT is applied, the performance of GPT-4 drops to 45.42, which is lower than the direct decoding result of 48.86.

From Table~\ref{tab:LLM_score}, we find the proposed MatPlotAgent can improve the plotting capabilities of several models. For GPT-4 and GPT-3.5, MatPlotAgent leads to significant improvements of 12.30 and 9.48, respectively. For the other five open-source LLMs, MatPlotAgent improves the performance of four models. With MatPlotAgent, the open-source Magicoder-S-DS-6.7B model even surpasses GPT-4 with direct decoding (51.70 vs. 48.86), showcasing the effectiveness of our method.

To investigate the generalizability of MatPlotAgent across various multi-modal LLMs, we present the results of employing Gemini Pro Vision as the visual agent in Table~\ref{tab:LLM_Gemini_vf_score}.
We observe considerable improvements of 7.87 and 5.45, respectively, over the direct decoding baseline. This evidence further demonstrates the model-agnostic characteristic of our approach, leveraging various multi-modal LLMs to achieve enhanced performance.

\begin{figure*}[t]
    \centering
    \includegraphics[width=0.99\linewidth]{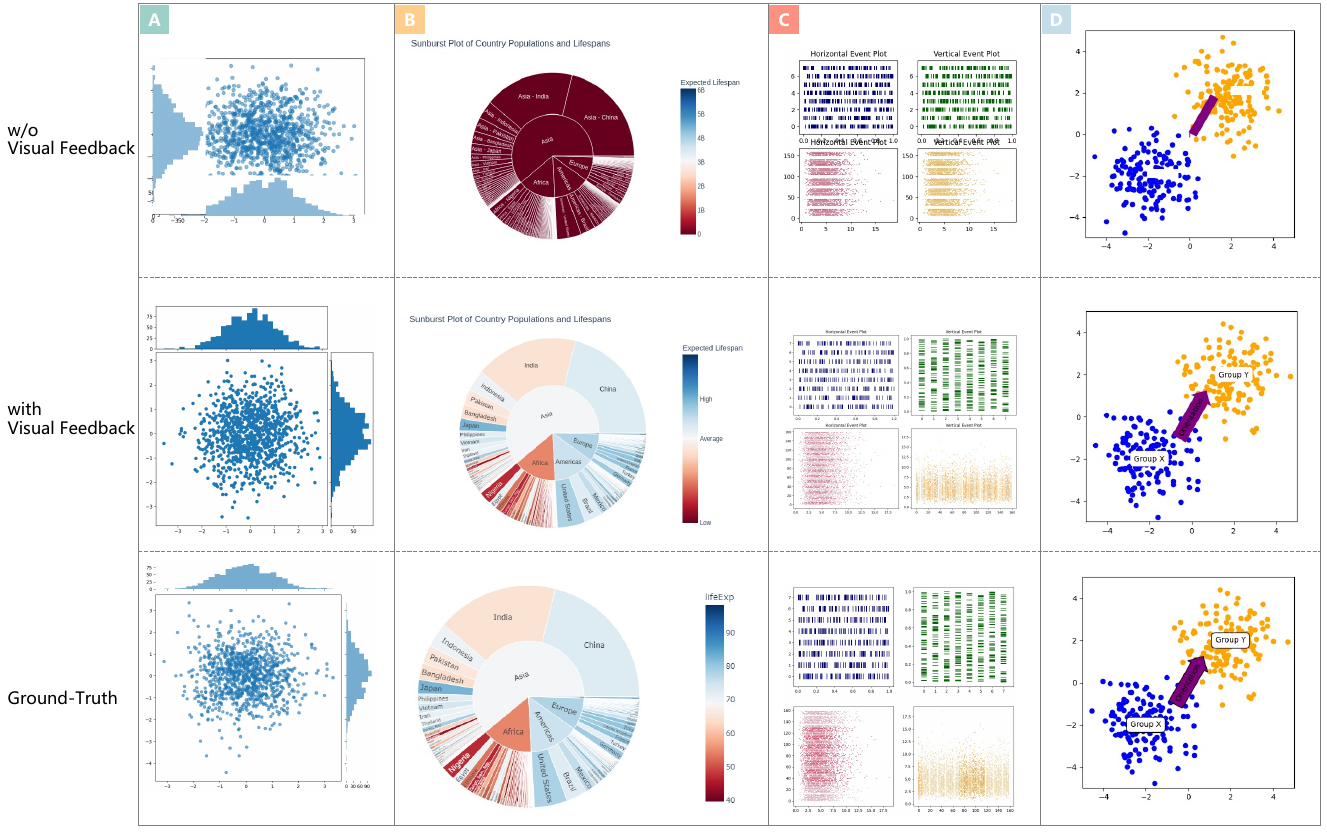}
    \caption{Examples to illustrate the effect of visual feedback. To investigate the effect of the visual feedback mechanism on different models, we display the outputs of two representative LLMs. Case A, B, and C are generated by GPT-4. Case D is generated by Magicoder-S-DS-6.7B.}
    \label{fig:ablation}
\end{figure*}

\subsection{Results on Qwen-Agent Code Interpreter Benchmark}
\label{sec:qwen_benchmark}
In Table~\ref{tab:qwen_benchmark_vis_study_results}, we detail the performance of MatPlotAgent on the visualization subset of the Qwen-Agent Code Interpreter Benchmark\footnote{\url{https://github.com/QwenLM/Qwen-Agent/tree/main/benchmark}}, which was recently published. According to their GitHub repository, GPT-4 achieved scores of 66.7 and 60.8 on the Visualization-Hard and Visualization-Easy subsets, respectively. Utilizing MatPlotAgent, we attained higher scores of 72.62 and 68.35 on these subsets.
When the visual feedback mechanism is disabled, MatPlotAgent reached scores of 66.67 and 65.82, reconfirming the necessity of visual feedback.

\subsection{Ablation Study}
\label{sec:ablation}
Compared to previous LLM-based coding agents~\cite{qian2023ChatDev,chen2024teaching}, the major contribution of the work lies in the newly proposed visual feedback mechanism, expected to leverage visual signals to enhance the quality of the output figure. To gain a deeper understanding of the impact of the visual feedback mechanism, we conduct both qualitative and quantitative analyses in this section.

Figure~\ref{fig:ablation} presents examples plotted by LLMs both with and without the visual feedback mechanism. We observe a clear improvement in the quality of the output figure with the visual feedback. 
For example, in case C, the text in the figure is jumbled, but this issue is resolved with the assistance of visual feedback.
It is important to note that the visual agent does not reference the ground-truth figure when generating feedback; it only examines the draft plotted by the model. 
Table~\ref{tab:ablation_study_results} also presents quantitative results of the visual feedback mechanism, indicating that the absence of visual feedback would result in significantly poorer outcomes for both GPT-4 and GPT-3.5. This reaffirms the importance of visual signals in the task of scientific data visualization.

\begin{table}[t]
\centering
\begin{tabular}{l cc}
\toprule
\textbf{Model} & \bf GPT-4 & \bf GPT-3.5 \\
\midrule
Direct Decod. & 48.86 & 38.03 \\
\cmidrule(lr){1-1} \cmidrule(lr){2-3}
MatPlotAgent & 61.16 & 47.51 \\
\ \ \ \ w/o Visual Feedback & 53.44 & 41.57 \\
\bottomrule
\end{tabular}
\caption{Effect of the visual feedback mechanism (GPT-4V visual agent).}
\label{tab:ablation_study_results}
\end{table}

\subsection{Case Study}

\begin{figure*}[t]
    \centering
    \includegraphics[width=0.99\linewidth]{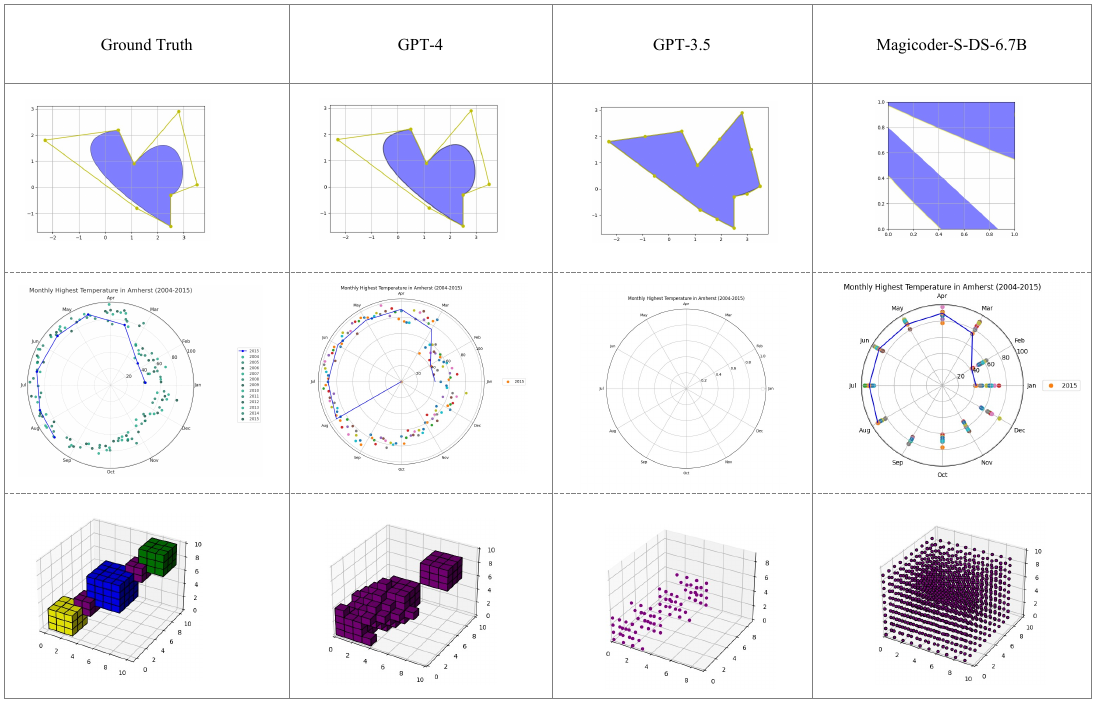}
    \caption{Case study of different models.}
    \label{fig:case_study}
\end{figure*}

We present output figures in Figure~\ref{fig:case_study}. The first example is relatively simple, correctly plotted by GPT-4 augmented with MatPlotAgent. The second example is more challenging; while GPT-4 and Magicoder-S-DS-6.7B can generate a draft, both omit some elements. The third example is the most difficult, where none of the three models can produce the correct result. 
These results indicate that the proposed MatPlotBench poses a significant challenge for current LLMs. Even the state-of-the-art LLM, GPT-4, equipped with MatPlotAgent, fails in some cases. We believe this benchmark will be effective not only for evaluating AI systems in scientific data visualization but also for assessing general capabilities such as coding and visual perception.

\section{Related Work}
\paragraph{Code LLMs}
Since the release of Codex~\cite{chen2021Codex}, many closed- and open-source code LLMs have been published, pushing the boundaries of LLMs’ capabilities to write functional code. Early open-source efforts include SantaCoder~\cite{allal2023santacoder} and StarCoder~\cite{li2023starcoder}.
More recently, the Code Llama~\cite{rozière2024code} series is released, including models of varying sizes.
DeepSeekCoder~\cite{guo2024deepseekcoder}, a series of open-source code models ranging in size from 1.3B to 33B, has also garnered significant attention for its impressive performance on general coding benchmarks.
\citet{wei2023magicoder} introduce a novel data augmentation method for automatically creating high-quality fine-tuning data. The resulting Magicoder model surpasses a wide array of open-source code LLMs in performance.

\paragraph{LLM Agents}
Recently, a wide range of LLM-based agent frameworks is proposed to explore LLMs’ potential in real-world scenarios~\cite{Nakano2021WebGPTBQ,yao_webshop_2022,qin-etal-2023-webcpm,zhou2023webarena}.
OpenAgents~\cite{xie2023openagents} proposed an open platform that leverages LLM agents in everyday situation by employing a Data Agent, a Plugins Agent, and a Web Agent.
\citet{Generative_agents} proposed an interactive simulation of human behavior in which software agents emulate realistic human actions and interactions through computation. Voyager~\cite{Wang2023VoyagerAO} introduced the fisrt LLM model-driven autonomous agent in Minecraft, designed to perpetually explore the environment, master various skills, and uncover new insights independently, without any human guidance.
ChatDev \cite{qian2023ChatDev} proposed creating a virtual, chat-driven software development enterprise that follows the traditional waterfall methodology.
In this study, we explore the capabilities of LLM-based agents in the task of scientific data visualization, a critical and practical area for contemporary researchers.

\section{Conclusion}

We propose to assess and enhance the capabilities of modern LLMs for scientific data visualization, a multifaceted task demanding coding and visual skills. We begin with the creation of MatPlotBench, a rigorous benchmark supporting automated quantitative evaluation that strongly aligns with human assessment. Additionally, we introduce MatPlotAgent, a model-agnostic mechanism employing visual feedback to enhance LLMs' plotting abilities. Experimental results demonstrate that MatPlotAgent enhances the performance of various LLMs.

\section{Limitations}

In this paper, we introduce MatPlotBench, a benchmark designed for scientific data visualization. However, the demands of scientific data visualization can vary significantly across disciplines. Since MatPlotBench is developed for general scientific data visualization, it may not encompass all domain-specific requirements, potentially restricting its applicability to certain fields. In the future, the data construction and evaluation approaches can be customized for specific domains if necessary.

\bibliography{custom}

\begin{thebibliography}{42}
\expandafter\ifx\csname natexlab\endcsname\relax\def\natexlab#1{#1}\fi

\bibitem[{Allal et~al.(2023)Allal, Li, Kocetkov, Mou, Akiki, Ferrandis, Muennighoff, Mishra, Gu, Dey, Umapathi, Anderson, Zi, Poirier, Schoelkopf, Troshin, Abulkhanov, Romero, Lappert, Toni, del Río, Liu, Bose, Bhattacharyya, Zhuo, Yu, Villegas, Zocca, Mangrulkar, Lansky, Nguyen, Contractor, Villa, Li, Bahdanau, Jernite, Hughes, Fried, Guha, de~Vries, and von Werra}]{allal2023santacoder}
Loubna~Ben Allal, Raymond Li, Denis Kocetkov, Chenghao Mou, Christopher Akiki, Carlos~Munoz Ferrandis, Niklas Muennighoff, Mayank Mishra, Alex Gu, Manan Dey, Logesh~Kumar Umapathi, Carolyn~Jane Anderson, Yangtian Zi, Joel~Lamy Poirier, Hailey Schoelkopf, Sergey Troshin, Dmitry Abulkhanov, Manuel Romero, Michael Lappert, Francesco~De Toni, Bernardo~García del Río, Qian Liu, Shamik Bose, Urvashi Bhattacharyya, Terry~Yue Zhuo, Ian Yu, Paulo Villegas, Marco Zocca, Sourab Mangrulkar, David Lansky, Huu Nguyen, Danish Contractor, Luis Villa, Jia Li, Dzmitry Bahdanau, Yacine Jernite, Sean Hughes, Daniel Fried, Arjun Guha, Harm de~Vries, and Leandro von Werra. 2023.
\newblock \href {http://arxiv.org/abs/2301.03988} {Santacoder: don't reach for the stars!}

\bibitem[{Azerbayev et~al.(2024)Azerbayev, Schoelkopf, Paster, Santos, McAleer, Jiang, Deng, Biderman, and Welleck}]{azerbayev2024llemma}
Zhangir Azerbayev, Hailey Schoelkopf, Keiran Paster, Marco~Dos Santos, Stephen~Marcus McAleer, Albert~Q. Jiang, Jia Deng, Stella Biderman, and Sean Welleck. 2024.
\newblock \href {https://openreview.net/forum?id=4WnqRR915j} {Llemma: An open language model for mathematics}.
\newblock In \emph{The Twelfth International Conference on Learning Representations}.

\bibitem[{Chen et~al.(2021)Chen, Tworek, Jun, Yuan, de~Oliveira~Pinto, Kaplan, Edwards, Burda, Joseph, Brockman, Ray, Puri, Krueger, Petrov, Khlaaf, Sastry, Mishkin, Chan, Gray, Ryder, Pavlov, Power, Kaiser, Bavarian, Winter, Tillet, Such, Cummings, Plappert, Chantzis, Barnes, Herbert-Voss, Guss, Nichol, Paino, Tezak, Tang, Babuschkin, Balaji, Jain, Saunders, Hesse, Carr, Leike, Achiam, Misra, Morikawa, Radford, Knight, Brundage, Murati, Mayer, Welinder, McGrew, Amodei, McCandlish, Sutskever, and Zaremba}]{chen2021Codex}
Mark Chen, Jerry Tworek, Heewoo Jun, Qiming Yuan, Henrique~Ponde de~Oliveira~Pinto, Jared Kaplan, Harri Edwards, Yuri Burda, Nicholas Joseph, Greg Brockman, Alex Ray, Raul Puri, Gretchen Krueger, Michael Petrov, Heidy Khlaaf, Girish Sastry, Pamela Mishkin, Brooke Chan, Scott Gray, Nick Ryder, Mikhail Pavlov, Alethea Power, Lukasz Kaiser, Mohammad Bavarian, Clemens Winter, Philippe Tillet, Felipe~Petroski Such, Dave Cummings, Matthias Plappert, Fotios Chantzis, Elizabeth Barnes, Ariel Herbert-Voss, William~Hebgen Guss, Alex Nichol, Alex Paino, Nikolas Tezak, Jie Tang, Igor Babuschkin, Suchir Balaji, Shantanu Jain, William Saunders, Christopher Hesse, Andrew~N. Carr, Jan Leike, Josh Achiam, Vedant Misra, Evan Morikawa, Alec Radford, Matthew Knight, Miles Brundage, Mira Murati, Katie Mayer, Peter Welinder, Bob McGrew, Dario Amodei, Sam McCandlish, Ilya Sutskever, and Wojciech Zaremba. 2021.
\newblock \href {http://arxiv.org/abs/2107.03374} {Evaluating large language models trained on code}.

\bibitem[{Chen et~al.(2024{\natexlab{a}})Chen, Su, Zuo, Yang, Yuan, Chan, Yu, Lu, Hung, Qian, Qin, Cong, Xie, Liu, Sun, and Zhou}]{chen2024agentverse}
Weize Chen, Yusheng Su, Jingwei Zuo, Cheng Yang, Chenfei Yuan, Chi-Min Chan, Heyang Yu, Yaxi Lu, Yi-Hsin Hung, Chen Qian, Yujia Qin, Xin Cong, Ruobing Xie, Zhiyuan Liu, Maosong Sun, and Jie Zhou. 2024{\natexlab{a}}.
\newblock \href {https://openreview.net/forum?id=EHg5GDnyq1} {Agentverse: Facilitating multi-agent collaboration and exploring emergent behaviors}.
\newblock In \emph{The Twelfth International Conference on Learning Representations}.

\bibitem[{Chen et~al.(2024{\natexlab{b}})Chen, Lin, Sch{\"a}rli, and Zhou}]{chen2024teaching}
Xinyun Chen, Maxwell Lin, Nathanael Sch{\"a}rli, and Denny Zhou. 2024{\natexlab{b}}.
\newblock \href {https://openreview.net/forum?id=KuPixIqPiq} {Teaching large language models to self-debug}.
\newblock In \emph{The Twelfth International Conference on Learning Representations}.

\bibitem[{Deng et~al.(2023)Deng, Gu, Zheng, Chen, Stevens, Wang, Sun, and Su}]{deng2023mindweb}
Xiang Deng, Yu~Gu, Boyuan Zheng, Shijie Chen, Samuel Stevens, Boshi Wang, Huan Sun, and Yu~Su. 2023.
\newblock \href {https://openreview.net/forum?id=kiYqbO3wqw} {Mind2web: Towards a generalist agent for the web}.
\newblock In \emph{Thirty-seventh Conference on Neural Information Processing Systems Datasets and Benchmarks Track}.

\bibitem[{Google(2023)}]{geminiteam2023gemini}
Gemini~Team Google. 2023.
\newblock \href {http://arxiv.org/abs/2312.11805} {Gemini: A family of highly capable multimodal models}.

\bibitem[{Guo et~al.(2024)Guo, Zhu, Yang, Xie, Dong, Zhang, Chen, Bi, Wu, Li, Luo, Xiong, and Liang}]{guo2024deepseekcoder}
Daya Guo, Qihao Zhu, Dejian Yang, Zhenda Xie, Kai Dong, Wentao Zhang, Guanting Chen, Xiao Bi, Y.~Wu, Y.~K. Li, Fuli Luo, Yingfei Xiong, and Wenfeng Liang. 2024.
\newblock \href {http://arxiv.org/abs/2401.14196} {Deepseek-coder: When the large language model meets programming -- the rise of code intelligence}.

\bibitem[{Kojima et~al.(2022{\natexlab{a}})Kojima, Gu, Reid, Matsuo, and Iwasawa}]{NEURIPS2022_zero-shot-cot}
Takeshi Kojima, Shixiang~(Shane) Gu, Machel Reid, Yutaka Matsuo, and Yusuke Iwasawa. 2022{\natexlab{a}}.
\newblock Large language models are zero-shot reasoners.
\newblock In \emph{Advances in Neural Information Processing Systems}, volume~35, pages 22199--22213.

\bibitem[{Kojima et~al.(2022{\natexlab{b}})Kojima, Gu, Reid, Matsuo, and Iwasawa}]{NEURIPS2022_8bb0d291}
Takeshi Kojima, Shixiang~(Shane) Gu, Machel Reid, Yutaka Matsuo, and Yusuke Iwasawa. 2022{\natexlab{b}}.
\newblock \href {https://proceedings.neurips.cc/paper_files/paper/2022/file/8bb0d291acd4acf06ef112099c16f326-Paper-Conference.pdf} {Large language models are zero-shot reasoners}.
\newblock In \emph{Advances in Neural Information Processing Systems}.

\bibitem[{Kwon et~al.(2023)Kwon, Li, Zhuang, Sheng, Zheng, Yu, Gonzalez, Zhang, and Stoica}]{kwon2023efficient}
Woosuk Kwon, Zhuohan Li, Siyuan Zhuang, Ying Sheng, Lianmin Zheng, Cody~Hao Yu, Joseph~E. Gonzalez, Hao Zhang, and Ion Stoica. 2023.
\newblock Efficient memory management for large language model serving with pagedattention.
\newblock In \emph{Proceedings of the ACM SIGOPS 29th Symposium on Operating Systems Principles}.

\bibitem[{Lai et~al.(2023)Lai, Li, Wang, Zhang, Zhong, Zettlemoyer, Yih, Fried, Wang, and Yu}]{Lai2023DS1000}
Yuhang Lai, Chengxi Li, Yiming Wang, Tianyi Zhang, Ruiqi Zhong, Luke Zettlemoyer, Wen-Tau Yih, Daniel Fried, Sida Wang, and Tao Yu. 2023.
\newblock {DS}-1000: A natural and reliable benchmark for data science code generation.
\newblock In \emph{Proceedings of the 40th International Conference on Machine Learning}.

\bibitem[{Li et~al.(2023{\natexlab{a}})Li, Hammoud, Itani, Khizbullin, and Ghanem}]{li2023camel}
Guohao Li, Hasan Abed Al~Kader Hammoud, Hani Itani, Dmitrii Khizbullin, and Bernard Ghanem. 2023{\natexlab{a}}.
\newblock \href {https://openreview.net/forum?id=3IyL2XWDkG} {{CAMEL}: Communicative agents for ''mind'' exploration of large language model society}.
\newblock In \emph{Thirty-seventh Conference on Neural Information Processing Systems}.

\bibitem[{Li et~al.(2023{\natexlab{b}})Li, allal, Zi, Muennighoff, Kocetkov, Mou, Marone, Akiki, LI, Chim, Liu, Zheltonozhskii, Zhuo, Wang, Dehaene, Lamy-Poirier, Monteiro, Gontier, Yee, Umapathi, Zhu, Lipkin, Oblokulov, Wang, Murthy, Stillerman, Patel, Abulkhanov, Zocca, Dey, Zhang, Bhattacharyya, Yu, Luccioni, Villegas, Zhdanov, Lee, Timor, Ding, Schlesinger, Schoelkopf, Ebert, Dao, Mishra, Gu, Anderson, Dolan-Gavitt, Contractor, Reddy, Fried, Bahdanau, Jernite, Ferrandis, Hughes, Wolf, Guha, Werra, and de~Vries}]{li2023starcoder}
Raymond Li, Loubna~Ben allal, Yangtian Zi, Niklas Muennighoff, Denis Kocetkov, Chenghao Mou, Marc Marone, Christopher Akiki, Jia LI, Jenny Chim, Qian Liu, Evgenii Zheltonozhskii, Terry~Yue Zhuo, Thomas Wang, Olivier Dehaene, Joel Lamy-Poirier, Joao Monteiro, Nicolas Gontier, Ming-Ho Yee, Logesh~Kumar Umapathi, Jian Zhu, Ben Lipkin, Muhtasham Oblokulov, Zhiruo Wang, Rudra Murthy, Jason~T Stillerman, Siva~Sankalp Patel, Dmitry Abulkhanov, Marco Zocca, Manan Dey, Zhihan Zhang, Urvashi Bhattacharyya, Wenhao Yu, Sasha Luccioni, Paulo Villegas, Fedor Zhdanov, Tony Lee, Nadav Timor, Jennifer Ding, Claire~S Schlesinger, Hailey Schoelkopf, Jan Ebert, Tri Dao, Mayank Mishra, Alex Gu, Carolyn~Jane Anderson, Brendan Dolan-Gavitt, Danish Contractor, Siva Reddy, Daniel Fried, Dzmitry Bahdanau, Yacine Jernite, Carlos~Mu{\~n}oz Ferrandis, Sean Hughes, Thomas Wolf, Arjun Guha, Leandro~Von Werra, and Harm de~Vries. 2023{\natexlab{b}}.
\newblock \href {https://openreview.net/forum?id=KoFOg41haE} {Starcoder: may the source be with you!}
\newblock \emph{Transactions on Machine Learning Research}.
\newblock Reproducibility Certification.

\bibitem[{Liu et~al.(2024)Liu, Yu, Zhang, Xu, Lei, Lai, Gu, Ding, Men, Yang, Zhang, Deng, Zeng, Du, Zhang, Shen, Zhang, Su, Sun, Huang, Dong, and Tang}]{liu2024agentbench}
Xiao Liu, Hao Yu, Hanchen Zhang, Yifan Xu, Xuanyu Lei, Hanyu Lai, Yu~Gu, Hangliang Ding, Kaiwen Men, Kejuan Yang, Shudan Zhang, Xiang Deng, Aohan Zeng, Zhengxiao Du, Chenhui Zhang, Sheng Shen, Tianjun Zhang, Yu~Su, Huan Sun, Minlie Huang, Yuxiao Dong, and Jie Tang. 2024.
\newblock \href {https://openreview.net/forum?id=zAdUB0aCTQ} {Agentbench: Evaluating {LLM}s as agents}.
\newblock In \emph{The Twelfth International Conference on Learning Representations}.

\bibitem[{Lu et~al.(2023)Lu, Peng, Cheng, Galley, Chang, Wu, Zhu, and Gao}]{lu2023chameleon}
Pan Lu, Baolin Peng, Hao Cheng, Michel Galley, Kai-Wei Chang, Ying~Nian Wu, Song-Chun Zhu, and Jianfeng Gao. 2023.
\newblock \href {https://openreview.net/forum?id=HtqnVSCj3q} {Chameleon: Plug-and-play compositional reasoning with large language models}.
\newblock In \emph{Thirty-seventh Conference on Neural Information Processing Systems}.

\bibitem[{Luo et~al.(2023{\natexlab{a}})Luo, Sun, Xu, Zhao, Lou, Tao, Geng, Lin, Chen, and Zhang}]{luo2023wizardmath}
Haipeng Luo, Qingfeng Sun, Can Xu, Pu~Zhao, Jianguang Lou, Chongyang Tao, Xiubo Geng, Qingwei Lin, Shifeng Chen, and Dongmei Zhang. 2023{\natexlab{a}}.
\newblock Wizardmath: Empowering mathematical reasoning for large language models via reinforced evol-instruct.
\newblock \emph{arXiv preprint arXiv:2308.09583}.

\bibitem[{Luo et~al.(2023{\natexlab{b}})Luo, Xu, Zhao, Sun, Geng, Hu, Tao, Ma, Lin, and Jiang}]{luo2023wizardcoder}
Ziyang Luo, Can Xu, Pu~Zhao, Qingfeng Sun, Xiubo Geng, Wenxiang Hu, Chongyang Tao, Jing Ma, Qingwei Lin, and Daxin Jiang. 2023{\natexlab{b}}.
\newblock \href {http://arxiv.org/abs/2306.08568} {Wizardcoder: Empowering code large language models with evol-instruct}.

\bibitem[{Nakano et~al.(2021)Nakano, Hilton, Balaji, Wu, Long, Kim, Hesse, Jain, Kosaraju, Saunders, Jiang, Cobbe, Eloundou, Krueger, Button, Knight, Chess, and Schulman}]{Nakano2021WebGPTBQ}
Reiichiro Nakano, Jacob Hilton, Suchir Balaji, Jeff Wu, Ouyang Long, Christina Kim, Christopher Hesse, Shantanu Jain, Vineet Kosaraju, William Saunders, Xu~Jiang, Karl Cobbe, Tyna Eloundou, Gretchen Krueger, Kevin Button, Matthew Knight, Benjamin Chess, and John Schulman. 2021.
\newblock \href {https://api.semanticscholar.org/CorpusID:245329531} {Webgpt: Browser-assisted question-answering with human feedback}.
\newblock \emph{ArXiv}, abs/2112.09332.

\bibitem[{OpenAI(2023)}]{Achiam2023GPT4TR}
OpenAI. 2023.
\newblock \href {https://api.semanticscholar.org/CorpusID:257532815} {Gpt-4 technical report}.

\bibitem[{Park et~al.(2023)Park, O'Brien, Cai, Morris, Liang, and Bernstein}]{Generative_agents}
Joon~Sung Park, Joseph O'Brien, Carrie~Jun Cai, Meredith~Ringel Morris, Percy Liang, and Michael~S. Bernstein. 2023.
\newblock \href {https://doi.org/10.1145/3586183.3606763} {Generative agents: Interactive simulacra of human behavior}.
\newblock In \emph{Proceedings of the 36th Annual ACM Symposium on User Interface Software and Technology}, UIST '23, New York, NY, USA. Association for Computing Machinery.

\bibitem[{Qian et~al.(2023{\natexlab{a}})Qian, Cong, Liu, Yang, Chen, Su, Dang, Li, Xu, Li, Liu, and Sun}]{qian2023ChatDev}
Chen Qian, Xin Cong, Wei Liu, Cheng Yang, Weize Chen, Yusheng Su, Yufan Dang, Jiahao Li, Juyuan Xu, Dahai Li, Zhiyuan Liu, and Maosong Sun. 2023{\natexlab{a}}.
\newblock \href {http://arxiv.org/abs/2307.07924} {Communicative agents for software development}.

\bibitem[{Qian et~al.(2023{\natexlab{b}})Qian, Han, Fung, Qin, Liu, and Ji}]{qian-etal-2023-creator}
Cheng Qian, Chi Han, Yi~Fung, Yujia Qin, Zhiyuan Liu, and Heng Ji. 2023{\natexlab{b}}.
\newblock \href {https://doi.org/10.18653/v1/2023.findings-emnlp.462} {{CREATOR}: Tool creation for disentangling abstract and concrete reasoning of large language models}.
\newblock In \emph{Findings of the Association for Computational Linguistics: EMNLP 2023}, pages 6922--6939, Singapore. Association for Computational Linguistics.

\bibitem[{Qin et~al.(2023)Qin, Cai, Jin, Yan, Liang, Zhu, Lin, Han, Ding, Wang, Xie, Qi, Liu, Sun, and Zhou}]{qin-etal-2023-webcpm}
Yujia Qin, Zihan Cai, Dian Jin, Lan Yan, Shihao Liang, Kunlun Zhu, Yankai Lin, Xu~Han, Ning Ding, Huadong Wang, Ruobing Xie, Fanchao Qi, Zhiyuan Liu, Maosong Sun, and Jie Zhou. 2023.
\newblock \href {https://doi.org/10.18653/v1/2023.acl-long.499} {{W}eb{CPM}: Interactive web search for {C}hinese long-form question answering}.
\newblock In \emph{Proceedings of the 61st Annual Meeting of the Association for Computational Linguistics (Volume 1: Long Papers)}, pages 8968--8988, Toronto, Canada. Association for Computational Linguistics.

\bibitem[{Qin et~al.(2024)Qin, Liang, Ye, Zhu, Yan, Lu, Lin, Cong, Tang, Qian, Zhao, Hong, Tian, Xie, Zhou, Gerstein, dahai li, Liu, and Sun}]{qin2024toolllm}
Yujia Qin, Shihao Liang, Yining Ye, Kunlun Zhu, Lan Yan, Yaxi Lu, Yankai Lin, Xin Cong, Xiangru Tang, Bill Qian, Sihan Zhao, Lauren Hong, Runchu Tian, Ruobing Xie, Jie Zhou, Mark Gerstein, dahai li, Zhiyuan Liu, and Maosong Sun. 2024.
\newblock \href {https://openreview.net/forum?id=dHng2O0Jjr} {Tool{LLM}: Facilitating large language models to master 16000+ real-world {API}s}.
\newblock In \emph{The Twelfth International Conference on Learning Representations}.

\bibitem[{Ramesh et~al.(2021)Ramesh, Pavlov, Goh, Gray, Voss, Radford, Chen, and Sutskever}]{ramesh2021zeroshotDalle}
Aditya Ramesh, Mikhail Pavlov, Gabriel Goh, Scott Gray, Chelsea Voss, Alec Radford, Mark Chen, and Ilya Sutskever. 2021.
\newblock \href {http://arxiv.org/abs/2102.12092} {Zero-shot text-to-image generation}.

\bibitem[{Rombach et~al.(2022)Rombach, Blattmann, Lorenz, Esser, and Ommer}]{rombach2022highresolutionStableDiffusion}
Robin Rombach, Andreas Blattmann, Dominik Lorenz, Patrick Esser, and Björn Ommer. 2022.
\newblock \href {http://arxiv.org/abs/2112.10752} {High-resolution image synthesis with latent diffusion models}.

\bibitem[{Rozière et~al.(2024)Rozière, Gehring, Gloeckle, Sootla, Gat, Tan, Adi, Liu, Sauvestre, Remez, Rapin, Kozhevnikov, Evtimov, Bitton, Bhatt, Ferrer, Grattafiori, Xiong, Défossez, Copet, Azhar, Touvron, Martin, Usunier, Scialom, and Synnaeve}]{rozière2024code}
Baptiste Rozière, Jonas Gehring, Fabian Gloeckle, Sten Sootla, Itai Gat, Xiaoqing~Ellen Tan, Yossi Adi, Jingyu Liu, Romain Sauvestre, Tal Remez, Jérémy Rapin, Artyom Kozhevnikov, Ivan Evtimov, Joanna Bitton, Manish Bhatt, Cristian~Canton Ferrer, Aaron Grattafiori, Wenhan Xiong, Alexandre Défossez, Jade Copet, Faisal Azhar, Hugo Touvron, Louis Martin, Nicolas Usunier, Thomas Scialom, and Gabriel Synnaeve. 2024.
\newblock \href {http://arxiv.org/abs/2308.12950} {Code llama: Open foundation models for code}.

\bibitem[{Saharia et~al.(2022)Saharia, Chan, Saxena, Li, Whang, Denton, Ghasemipour, Ayan, Mahdavi, Lopes, Salimans, Ho, Fleet, and Norouzi}]{saharia2022photorealisticImagen}
Chitwan Saharia, William Chan, Saurabh Saxena, Lala Li, Jay Whang, Emily Denton, Seyed Kamyar~Seyed Ghasemipour, Burcu~Karagol Ayan, S.~Sara Mahdavi, Rapha~Gontijo Lopes, Tim Salimans, Jonathan Ho, David~J Fleet, and Mohammad Norouzi. 2022.
\newblock \href {http://arxiv.org/abs/2205.11487} {Photorealistic text-to-image diffusion models with deep language understanding}.

\bibitem[{Schick et~al.(2023)Schick, Dwivedi-Yu, Dessi, Raileanu, Lomeli, Hambro, Zettlemoyer, Cancedda, and Scialom}]{schick2023toolformer}
Timo Schick, Jane Dwivedi-Yu, Roberto Dessi, Roberta Raileanu, Maria Lomeli, Eric Hambro, Luke Zettlemoyer, Nicola Cancedda, and Thomas Scialom. 2023.
\newblock \href {https://openreview.net/forum?id=Yacmpz84TH} {Toolformer: Language models can teach themselves to use tools}.
\newblock In \emph{Thirty-seventh Conference on Neural Information Processing Systems}.

\bibitem[{Shao et~al.(2024)Shao, Wang, Zhu, Xu, Song, Zhang, Li, Wu, and Guo}]{shao2024deepseekmath}
Zhihong Shao, Peiyi Wang, Qihao Zhu, Runxin Xu, Junxiao Song, Mingchuan Zhang, Y.~K. Li, Y.~Wu, and Daya Guo. 2024.
\newblock \href {http://arxiv.org/abs/2402.03300} {Deepseekmath: Pushing the limits of mathematical reasoning in open language models}.

\bibitem[{Shinn et~al.(2023)Shinn, Cassano, Gopinath, Narasimhan, and Yao}]{shinn2023reflexion}
Noah Shinn, Federico Cassano, Ashwin Gopinath, Karthik~R Narasimhan, and Shunyu Yao. 2023.
\newblock \href {https://openreview.net/forum?id=vAElhFcKW6} {Reflexion: language agents with verbal reinforcement learning}.
\newblock In \emph{Thirty-seventh Conference on Neural Information Processing Systems}.

\bibitem[{Wang et~al.(2023)Wang, Xie, Jiang, Mandlekar, Xiao, Zhu, Fan, and Anandkumar}]{Wang2023VoyagerAO}
Guanzhi Wang, Yuqi Xie, Yunfan Jiang, Ajay Mandlekar, Chaowei Xiao, Yuke Zhu, Linxi~(Jim) Fan, and Anima Anandkumar. 2023.
\newblock \href {https://api.semanticscholar.org/CorpusID:258887849} {Voyager: An open-ended embodied agent with large language models}.
\newblock \emph{ArXiv}, abs/2305.16291.

\bibitem[{Wei et~al.(2022)Wei, Wang, Schuurmans, Bosma, ichter, Xia, Chi, Le, and Zhou}]{NEURIPS2022_9d560961_CoT}
Jason Wei, Xuezhi Wang, Dale Schuurmans, Maarten Bosma, brian ichter, Fei Xia, Ed~Chi, Quoc~V Le, and Denny Zhou. 2022.
\newblock \href {https://proceedings.neurips.cc/paper_files/paper/2022/file/9d5609613524ecf4f15af0f7b31abca4-Paper-Conference.pdf} {Chain-of-thought prompting elicits reasoning in large language models}.
\newblock In \emph{Advances in Neural Information Processing Systems}, volume~35, pages 24824--24837. Curran Associates, Inc.

\bibitem[{Wei et~al.(2023)Wei, Wang, Liu, Ding, and Zhang}]{wei2023magicoder}
Yuxiang Wei, Zhe Wang, Jiawei Liu, Yifeng Ding, and Lingming Zhang. 2023.
\newblock \href {http://arxiv.org/abs/2312.02120} {Magicoder: Source code is all you need}.

\bibitem[{Xie et~al.(2023)Xie, Zhou, Cheng, Shi, Weng, Liu, Hua, Zhao, Liu, Liu, Liu, Xu, Su, Shin, Xiong, and Yu}]{xie2023openagents}
Tianbao Xie, Fan Zhou, Zhoujun Cheng, Peng Shi, Luoxuan Weng, Yitao Liu, Toh~Jing Hua, Junning Zhao, Qian Liu, Che Liu, Leo~Z. Liu, Yiheng Xu, Hongjin Su, Dongchan Shin, Caiming Xiong, and Tao Yu. 2023.
\newblock \href {http://arxiv.org/abs/2310.10634} {Openagents: An open platform for language agents in the wild}.

\bibitem[{Xu et~al.(2023)Xu, Wang, Li, Luo, Wang, Liu, and Liu}]{xu2023exploringWerewolf}
Yuzhuang Xu, Shuo Wang, Peng Li, Fuwen Luo, Xiaolong Wang, Weidong Liu, and Yang Liu. 2023.
\newblock \href {http://arxiv.org/abs/2309.04658} {Exploring large language models for communication games: An empirical study on werewolf}.

\bibitem[{Yao et~al.(2022)Yao, Chen, Yang, and Narasimhan}]{yao_webshop_2022}
Shunyu Yao, Howard Chen, John Yang, and Karthik Narasimhan. 2022.
\newblock \href {https://proceedings.neurips.cc/paper_files/paper/2022/file/82ad13ec01f9fe44c01cb91814fd7b8c-Paper-Conference.pdf} {{WebShop}: {Towards} {Scalable} {Real}-{World} {Web} {Interaction} with {Grounded} {Language} {Agents}}.
\newblock In \emph{Advances in {Neural} {Information} {Processing} {Systems}}, volume~35, pages 20744--20757. Curran Associates, Inc.

\bibitem[{Yao et~al.(2023{\natexlab{a}})Yao, Yu, Zhao, Shafran, Griffiths, Cao, and Narasimhan}]{yao2023treeToT}
Shunyu Yao, Dian Yu, Jeffrey Zhao, Izhak Shafran, Thomas~L. Griffiths, Yuan Cao, and Karthik~R Narasimhan. 2023{\natexlab{a}}.
\newblock \href {https://openreview.net/forum?id=5Xc1ecxO1h} {Tree of thoughts: Deliberate problem solving with large language models}.
\newblock In \emph{Thirty-seventh Conference on Neural Information Processing Systems}.

\bibitem[{Yao et~al.(2023{\natexlab{b}})Yao, Zhao, Yu, Du, Shafran, Narasimhan, and Cao}]{yao2023react}
Shunyu Yao, Jeffrey Zhao, Dian Yu, Nan Du, Izhak Shafran, Karthik~R Narasimhan, and Yuan Cao. 2023{\natexlab{b}}.
\newblock \href {https://openreview.net/forum?id=WE_vluYUL-X} {React: Synergizing reasoning and acting in language models}.
\newblock In \emph{The Eleventh International Conference on Learning Representations}.

\bibitem[{Yu et~al.(2024)Yu, Jiang, Shi, YU, Liu, Zhang, Kwok, Li, Weller, and Liu}]{yu2024metamath}
Longhui Yu, Weisen Jiang, Han Shi, Jincheng YU, Zhengying Liu, Yu~Zhang, James Kwok, Zhenguo Li, Adrian Weller, and Weiyang Liu. 2024.
\newblock \href {https://openreview.net/forum?id=N8N0hgNDRt} {Metamath: Bootstrap your own mathematical questions for large language models}.
\newblock In \emph{The Twelfth International Conference on Learning Representations}.

\bibitem[{Zhou et~al.(2023)Zhou, Xu, Zhu, Zhou, Lo, Sridhar, Cheng, Ou, Bisk, Fried, Alon, and Neubig}]{zhou2023webarena}
Shuyan Zhou, Frank~F. Xu, Hao Zhu, Xuhui Zhou, Robert Lo, Abishek Sridhar, Xianyi Cheng, Tianyue Ou, Yonatan Bisk, Daniel Fried, Uri Alon, and Graham Neubig. 2023.
\newblock \href {https://openreview.net/forum?id=rmiwIL98uQ} {Webarena: A realistic web environment for building autonomous agents}.
\newblock In \emph{Second Agent Learning in Open-Endedness Workshop}.

\end{thebibliography}

\appendix

\section{Detailed Prompts}
\label{sec:appendix}

To better understand MatPlotBench and MatPlotAgent, we list the prompts for automatic evaluation and the three modules in MatPlotAgent, including the query expansion module, the code agent, and the visual agent.

\subsection{Evaluation Prompts}

The automatic evaluation prompt primarily requires GPT-4V to provide a score between 0 and 100 for the model-generated plot, with reference to the ground truth plot.

\begin{figure}[h]
\begin{tcolorbox}[colback=blue!2,colframe=blue!50!black]
\small
\texttt{You are an excellent judge at evaluating visualization plots between a model-generated plot and the ground truth. You will be giving scores on how well it matches the ground truth plot. \vspace{4pt} \\
The generated plot will be given to you as the first figure. If the first figure is blank, that means the code failed to generate a figure. \vspace{4pt} \\
Another plot will be given to you as the second figure, which is the desired outcome of the user query, meaning it is the ground truth for you to reference. \vspace{4pt} \\
Please compare the two figures head to head and rate them.
Suppose the second figure has a score of 100, rate the first figure on a scale from 0 to 100. \vspace{4pt} \\
Scoring should be carried out regarding the plot correctness: 
Compare closely between the generated plot and the ground truth, the more resemblance the generated plot has compared to the ground truth, the higher the score. The score should be proportionate to the resemblance between the two plots. \vspace{4pt} \\
In some rare occurrences, see if the data points are generated randomly according to the query, if so, the generated plot may not perfectly match the ground truth, but it is correct nonetheless. \vspace{4pt} \\
Only rate the first figure, the second figure is only for reference. \vspace{4pt} \\
If the first figure is blank, that means the code failed to generate a figure. Give a score of 0 on the Plot correctness. \vspace{4pt} \\
After scoring from the above aspect, please give a final score. The final score is preceded by the [FINAL SCORE] token. \vspace{4pt} \\
For example [FINAL SCORE]: 40.}
\end{tcolorbox}
\caption{Automatic evaluation prompt for GPT-4V.}
\label{fig:4v_prompt}
\end{figure}

\subsection{Prompts for MatPlotAgent}

The query expansion prompt mainly requires LLMs to generate step-by-step, detailed instructions on how to use Python code to fulfill the requirements specified by users, as shown in Figure~\ref{fig:Query_Expansion_prompt}.

For the code agent, there are two prompts for the code generation process and the self-debugging mechanism. The code generation prompt mainly requires LLMs to generate executable code according to the user query to plot and save the output figure, as shown in Figure~\ref{fig:Code_Generation_prompt}. The self-debugging prompt mainly requires LLMs to correct the buggy code according to the error message from a Python interpreter, as displayed in Figure~\ref{fig:Self_Debugging_prompt}.

The visual agent prompt mainly requires multi-modal LLMs to firstly understand the user query and analyze the draft plot, then generate the visual feedback to refine the draft, as shown in Figure~\ref{fig:Visual_Agent_prompt}.

\begin{figure}[h]
\begin{tcolorbox}[colback=blue!2,colframe=blue!50!black]
\small
\textbf{SYSTEM PROMPT: } \texttt{According to the user query, expand and solidify the query into a step by step detailed instruction (or comment) on how to write python code to fulfill the user query's requirements. Import the appropriate libraries. Pinpoint the correct library functions to call and set each parameter in every function call accordingly.}
\vspace{8pt} \\
\textbf{USER PROMPT: } \texttt{Here is the user query: [User Query]:
"""
\{\{query\}\}
"""
You should understand what the query's requirements are, and output step by step, detailed instructions on how to use python code to fulfill these requirements. Include what libraries to import, what library functions to call, how to set the parameters in each function correctly, how to prepare the data, how to manipulate the data so that it becomes appropriate for later functions to call etc,. Make sure the code to be executable and correctly generate the desired output in the user query.}
\end{tcolorbox}
\caption{The query expansion prompt in MatPlotAgent.}
\label{fig:Query_Expansion_prompt}
\end{figure}

\begin{figure}[h]
\begin{tcolorbox}[colback=blue!2,colframe=blue!50!black]
\small
\textbf{SYSTEM PROMPT: } \texttt{You are a cutting-edge super capable code generation LLM. You will be given a natural language query, generate a runnable python code to satisfy all the requirements in the query. You can use any python library you want. When you complete a plot, remember to save it to a png file.}
\vspace{8pt} \\
\textbf{USER PROMPT: } \texttt{Here is the query:
"""
\{\{query\}\}
"""
If the query requires data manipulation from a csv file, process the data from the csv file and draw the plot in one piece of code. When you complete a plot, remember to save it to a png file. The file name should be """\{\{file\_name\}\}""".}
\end{tcolorbox}
\caption{The code generation prompt in MatPlotAgent.}
\label{fig:Code_Generation_prompt}
\end{figure}

\begin{figure}[h]
\begin{tcolorbox}[colback=blue!2,colframe=blue!50!black]
\small
\textbf{USER PROMPT: } \texttt{There are some errors in the code you gave:
\{\{error\_message\}\}
please correct the errors.
Then give the complete code and don't omit anything even though you have given it in the above code.}
\end{tcolorbox}
\caption{The self-debugging prompt in MatPlotAgent.}
\label{fig:Self_Debugging_prompt}
\end{figure}

\begin{figure}[h]
\begin{tcolorbox}[colback=blue!2,colframe=blue!50!black]
\small
\textbf{SYSTEM PROMPT: } \texttt{Given a user query and an image of the current plot, please determine whether the plot has faithfully followed the user query. Your task is to provide instruction to make sure the plot has strictly completed the requirements of the query. Please output a detailed step by step instruction on how to use python code to enhance the plot.}
\vspace{8pt} \\
\textbf{USER PROMPT: } \texttt{Here is the user query: [Query]:
"""
\{\{query\}\}
"""
Carefully read and analyze the user query to understand the specific requirements. Check if the plot aligns with the user query in terms of data selection, plot type, and any specific customization. Look at the provided image of the plot. Assess the plot type, the data it represents, labels, titles, colors, and any other visual elements. Compare these elements with the requirements specified in the user query. Note any differences between the user query requirements and the current plot. Based on the identified discrepancies, provide step-by-step instructions on how to modify the Python code to meet the user query requirements. Suggest improvements for better visualization practices, such as clarity, readability, and aesthetics, while ensuring the primary focus is on meeting the user's specified requirements.
Remember to save the plot to a png file. The file name should be """\{\{file\_name\}\}"""}
\end{tcolorbox}
\caption{Prompt for the visual agent.}
\label{fig:Visual_Agent_prompt}
\end{figure}

\section{Human Evaluation Details}

We engage human annotators from computer science departments at various universities via social media. They are compensated for their work at a rate slightly higher than the prevailing market rate. All human annotators involved are informed that the collected data will be used solely for academic research purposes, and their personal information will not be disclosed.

\subsection{Evaluation Guide for Human Annotators}

Figure~\ref{fig:evaluation-rubric} gives detailed instructions for human annotators when scoring the model-generated plots.

\begin{figure*}[ht]
  \begin{tcolorbox}[colback=red!2,colframe=red!50!black,title=Evaluation Guide]
  \textbf{Plot Correctness (0-100 points)}
  \begin{itemize}
    \item \textbf{Exact Match (90-100 points)}: The generated plot is nearly identical to the ground truth, with only minor, negligible differences.
    \item \textbf{High Resemblance (70-89 points)}: The generated plot closely resembles the ground truth with some small but noticeable differences in data representation or styling.
    \item \textbf{Moderate Resemblance (50-69 points)}: The generated plot has a moderate level of similarity to the ground truth, but there are several noticeable differences that impact the plot's accuracy or interpretation.
    \item \textbf{Low Resemblance (30-49 points)}: The generated plot shares some similarities with the ground truth but has significant differences that change the overall message or interpretation of the data.
    \item \textbf{Poor Match (10-29 points)}: The generated plot has very little in common with the ground truth, with major discrepancies in data representation.
    \item \textbf{No Resemblance (1-9 points)}: The generated plot is completely different from the ground truth, with no discernible similarities in data representation.
    \item \textbf{Failure to Generate (0 points)}: The first figure is blank, indicating a failure to generate any plot.
  \end{itemize}
  \textbf{Special Considerations}
  \begin{itemize}
    \item In cases where the generated plot includes random data points that are correct in the context of the query, the plot should be evaluated for its correctness based on the query's intent, not solely on its visual match to the ground truth.
  \end{itemize}

 \textbf{[FINAL SCORE]: XX}

  \end{tcolorbox}
  \caption{Evaluation guide for human annotators when scoring the model-generated plots.}
  \label{fig:evaluation-rubric}
\end{figure*}
\end{document}